\documentclass[10pt,twocolumn,letterpaper]{article}

\usepackage{cvpr}
\usepackage{times}
\usepackage{epsfig}
\usepackage{graphicx}
\usepackage{amsmath}
\usepackage{amssymb}
\usepackage{booktabs}
\usepackage{colortbl}
\usepackage{xcolor}

\definecolor{tablecolor}{rgb}{0.0,0.0,0.0}
\definecolor{cwblue1}{rgb}{0.27,0.427,0.623}
\definecolor{cwblue2}{rgb}{0.286,0.454,0.658}
\definecolor{cwblue3}{rgb}{0.733,0.811,0.905}

\newcommand{\myparagraph}[1]{\textbf{#1} --- }  

\newcommand{\bq}{\mbox{\boldmath$q$}}

\newcommand{\bc}{\mbox{\boldmath$c$}}

\newcommand{\bh}{\mbox{\boldmath$h$}}
\newcommand{\bm}{\mbox{\boldmath$m$}}

\newcommand{\bp}{\mbox{\boldmath$p$}}
\newcommand{\br}{\mbox{\boldmath$r$}}

\newcommand{\bv}{\mbox{\boldmath$v$}}

\newcommand{\bx}{\mbox{\boldmath$x$}}
\newcommand{\by}{\mbox{\boldmath$y$}}
\newcommand{\bz}{\mbox{\boldmath$z$}}
\newcommand{\bl}{\mbox{\boldmath$l$}}

\newcommand{\bD}{\mbox{\boldmath$D$}}

\newcommand{\bM}{\mbox{\boldmath$M$}}

\newcommand{\bV}{\mbox{\boldmath$V$}}
\newcommand{\bW}{\mbox{\boldmath$W$}}
\newcommand{\bX}{\mbox{\boldmath$X$}}

\newcommand{\bZ}{\mbox{\boldmath$Z$}}

\usepackage[pagebackref=true,breaklinks=true,letterpaper=true,colorlinks,bookmarks=false]{hyperref}

\cvprfinalcopy 

\def\cvprPaperID{182} 

\ifcvprfinal\fi
\begin{document}

\title{Glimpse Clouds: Human Activity Recognition from Unstructured Feature Points}

\author{Fabien Baradel\textsuperscript{1}, Christian Wolf\textsuperscript{1,2}, Julien Mille\textsuperscript{3}, Graham W. Taylor\textsuperscript{4,5} \vspace*{2mm}\\
\textsuperscript{1} Univ Lyon, INSA-Lyon, CNRS, LIRIS, F-69621, Villeurbanne, France\\
\textsuperscript{2} INRIA, CITI Laboratory, Villeurbanne, France\\
\textsuperscript{3} Laboratoire d'Informatique de l'Universit\'e{} de Tours, INSA Centre Val de Loire, 41034 Blois, France\\
\textsuperscript{4} School of Engineering, University of Guelph, Guelph, Ontario, Canada\\
\textsuperscript{5} Vector Institute, Toronto, Ontario, Canada\\
{\tt\small \{fabien.baradel,christian.wolf\}@liris.cnrs.fr}, 
{\tt\small julien.mille@insa-cvl.fr},
{\tt\small gwtaylor@uoguelph.ca}\\
\url{https://fabienbaradel.github.io/cvpr18_glimpseclouds/}
}

\maketitle

\begin{figure*}[h!] \centering
        \includegraphics[width=15cm]{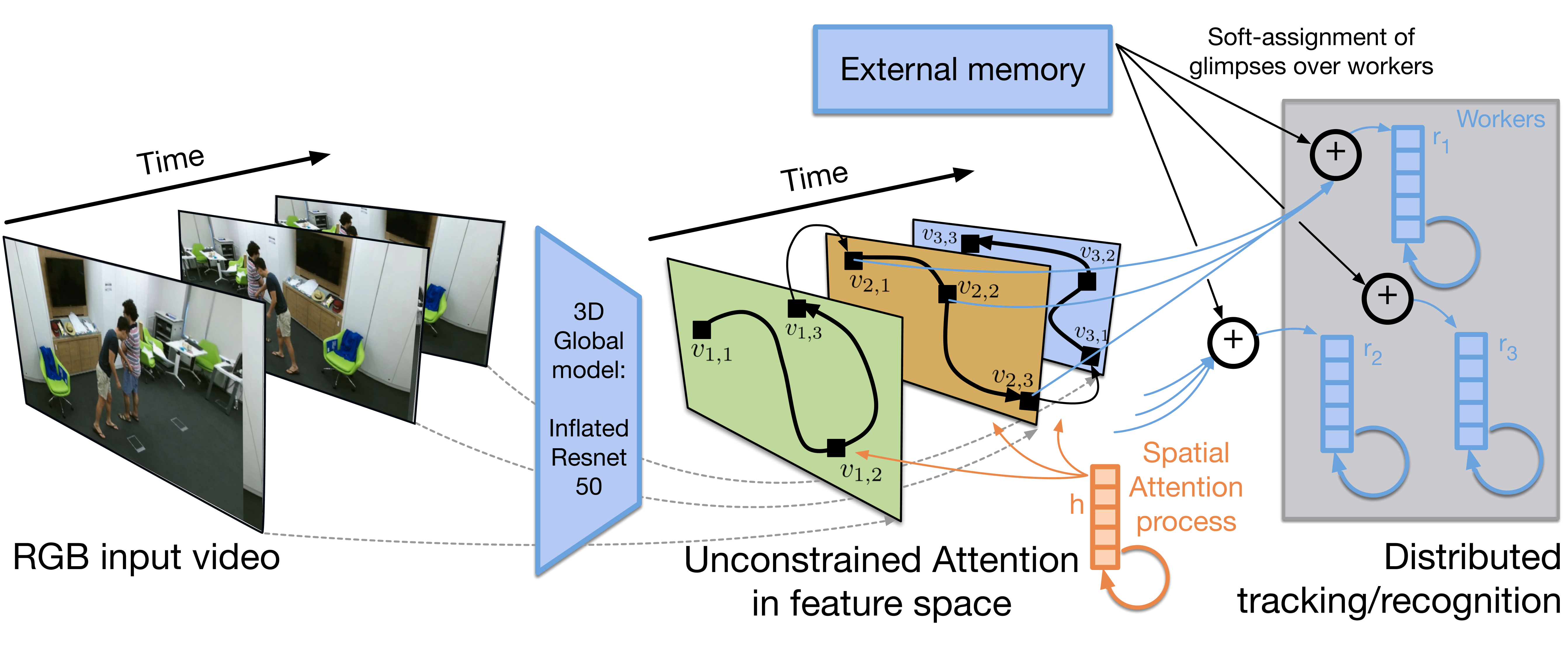}
    \caption{\label{fig:teaser}We recognize human activities from unstructured collections of spatio-temporal glimpses with distributed recurrent tracking/recognition and soft-assignment among glimpse points and trackers.
    }
\end{figure*}

\begin{abstract}
\noindent
We propose a method for human activity recognition from RGB data which does not rely on any pose information during test time, and which does not explicitly calculate pose information internally. Instead, a visual attention module learns to predict glimpse sequences in each frame. These glimpses correspond to interest points in the scene which are relevant to the classified activities. No spatial coherence is forced on the glimpse locations, which gives the module liberty to explore different points at each frame and better optimize the process of scrutinizing visual information.

Tracking and sequentially integrating this kind of unstructured data is a challenge, which we address by separating the set of glimpses from a set of recurrent 	 tracking/recognition workers.
These workers receive the glimpses, jointly performing subsequent motion tracking and prediction of the activity itself. The glimpses are soft-assigned to the workers, optimizing coherence of the assignments in space, time and feature space using an external memory module. No hard decisions are taken, i.e.~each glimpse point is assigned to all existing workers, albeit with different importance.
Our methods outperform state-of-the-art methods on the largest human activity recognition dataset available to-date; NTU RGB+D Dataset, and on a smaller human action recognition dataset Northwestern-UCLA Multiview Action 3D Dataset.
Our code is publicly available\footnote{\url{https://github.com/fabienbaradel/glimpse_clouds}}.
\end{abstract}

\section{Introduction}
\label{sec:introduction}

\noindent
We address the problem of human activity recognition in settings where activities are complex and diverse, performed by an individual or involving multiple people interacting. These activities may even include people interacting with objects or the environment. The usage of RGB-D cameras is very popular in this case, as it allows the use of articulated pose (skeletons) delivered in real time and relatively cheaply by some middleware. The exclusive usage of pose makes it possible to work on gesture and activity recognition without being a specialist in vision, and with significantly reduced dimensionality of the input data. The combined usage of pose and raw depth and/or RGB images can often boost performance over a solution that uses a single modality.

In this paper we propose a method which only uses raw RGB images at test time. We avoid the usage of articulated pose essentially for two reasons: (i) depth data is not always available, for example, in applications involving smaller or otherwise resource-constrained robots; and (ii) the question whether articulated pose is the optimal intermediate representation for activity recognition is unclear. We explore an alternative strategy, which consists of learning a local representation of the video through a visual attention process.

We conjecture that the replacement of articulated pose should keep one important property, which is its collection of local entities, which can be tracked over time and whose motion is relevant to the activity at hand. Instead of fixing the semantic meaning of these entities to the definition of a subset of the joints in the human body, we learn it discriminatively. In our strategy, the attention process is completely free to attend to arbitrary locations at each time instant. In particular, we do not impose any constraints on spatio-temporal coherence of glimpse locations, which allows the model to vary its focus within and across frames. Certain similarities can be made to human gaze patterns which saccade to different points in a scene.

Activities are highly correlated with motion, and therefore tracking the motion of specific points of visual interest is essential, yielding a distributed representation of the collection of glimpses. Appearance and motion features need to be collected over time from local points and integrated into a sequential decision model. However, tracking a set of glimpse points, whose location is not spatio-temporally smooth and whose semantic meaning can change from frame to frame, is a challenge. Our objective is to match new glimpses with past
ones of the same (or a nearby) location in the scene. Due to the unconstrained nature of the attention mechanism, we do not know when a point in the scene has been last scrutinized, or if it has been attended to in the past.

We solve this issue by separating the problem into two distinct parts: (i) selecting a distributed and local representation of $G$ glimpse points through a sequential recurrent attention model, and (ii) tracking the set of glimpses by a set of $C$ recurrent workers which sequentially integrate features, and participate in the final recognition of the activity (Fig.~\ref{fig:teaser}). In general, $G$ can be different from $C$, and the assignment between glimpses and workers is \emph{soft}. Each worker is potentially assigned to all glimpses, albeit to a varying degree. This assignment attention distribution is calculated with external memory based on regularities in space, time and feature space.

We summarize the main contributions of our paper as follows:
\begin{itemize}
\itemsep0em
\item We present a method for human activity recognition which does not require articulated pose during testing and which models activities two attentional processes; one extracting a set of glimpses per frame and one reasoning about entities over time.
\item This unstructured ``cloud'' of glimpses produced by the attention process are tracked over time using a set of trackers/recognizers, which are soft-assigned using external memory. Each tracker can potentially track multiple glimpses.
\item Articulated pose is used during \emph{training} time as an additional target, encouraging the attention process to focus on human structures.
\item All attentional mechanisms are executed in feature space which is calculated jointly with a global model processing the full input image.
\item We evaluate our method on the NTU RGB-D dataset, the largest available human activity dataset, where we outperform the state-of-the-art by a large margin.
\item We also show state-of-the-art results on a smaller human action recognition dataset: the Northwestern-UCLA Multiview Action 3D Dataset.
\end{itemize}

\section{Related Work}
\label{sec:relatedworks}

\noindent
\myparagraph{Activities, gestures and multimodal data}
Recent gesture and human activity recognition methods dealing with several modalities typically process 2D+T RGB and/or depth data as 3D. Sequences of frames are stacked into volumes and fed into convolutional layers at the first stages~\cite{Baccouche2011,Ji_PAMI2013, MolchanovYangCVPR2016, NeverovaWolfTaylorNeboutPAMI2016, WuPigouPAMI2016}. When additional pose data is available ~\cite{Luvizon_2018_CVPR}, the 3D joint positions are typically fed into a separate network. Preprocessing pose is reported to improve performance in some situations, e.g.~augmenting coordinates with velocities and acceleration~\cite{DBLP:conf/iccv/ZanfirLS13}.
Fusing pose and raw video modalities is traditionally done as late fusion~\cite{MolchanovYangCVPR2016}, or early through fusion layers~\cite{WuPigouPAMI2016}.

In contrast, our method does not require pose during testing and only leverages it during training for regularization.

\myparagraph{Recurrent architectures for action recognition}
Recurrent neural networks (or their variants) are employed in much of the contemporary work on activity recognition, and a recent trend is to make recurrent models local.
Part-aware LSTMs~\cite{Shahroudy2016} separate the memory cell of an LSTM network~\cite{Hochreiter1997} into part-based sub-cells and let the network learn long-term representations individually for each part, fusing the parts for output. Similarly, Du {\it et al}~\cite{Du_CVPR2015} use bi-directional LSTM layers which fit an anatomical hierarchy. Skeletons are split into anatomically-relevant parts (legs, arms, torso, {\it etc.}) and let subnetworks specialize on them. Lattice LSTMs partition the latent space over a grid which is aligned with the spatial input space \cite{Lattice_2017_ICCV}.

Our method, on the other hand, soft-assigns parts of the scene over multiple recurrent workers, where each worker can potentially integrate all points of the scene.





\myparagraph{Tracking and distributed recognition}
Structural RNNs \cite{JainStructuralRNN2016} bear a certain resemblance to our work. They handle the temporal evolution of tracked objects in videos with a set of RNNs, each of which correspond to cliques in a graph which models the spatio-temporal relationships between these objects. However, this graph is hand-crafted manually for each application, and the tracking of the objects is done using external trackers, which are not integrated into the neural model.

Our model, on the other hand, does not rely on external trackers and does not require the manual creation of a graph, since the assignments between objects (here, glimpses) and trackers are learned automatically.

\myparagraph{Attention mechanisms and external memory}
attention mechanisms focus selectively on parts of the scene which are the most relevant to the target task. Two classes of attention have emerged in recent years. \emph{Soft attention} weights each part of the observation dynamically~\cite{Bahdanau_ICLR2015, Kim_ICLR2017}. The objective function is usually differentiable, allowing gradient-based optimization. Soft attention was proposed for image \cite{ChoBengioMM2015, Xu_ICML2015} and video understanding ~\cite{Sharma2016a,Song2016,yeung2015every} with spatial, temporal and spatio-temporal variants.

Towards action recognition in particular, Sharma {\it et al}~\cite{Sharma2016a} proposed a recurrent mechanism from RGB data, which integrates convolutional features from different parts of a space-time volume. Song {\it et al} \cite{Song2016} propose separate spatial and temporal attention networks for action recognition from pose. At each frame, the spatial attention model gives more importance to the joints most relevant to the current action, whereas the temporal model selects frames.

On the other hand, \emph{hard attention}, which is the principle we adopt in this work, takes hard decisions when choosing parts of the input data. In a seminal paper, Mnih {\it et al}~\cite{Mnih_NIPS2014} proposed visual hard attention for image classification built around an RNN. The model selects the next location to focus on, based on past information. Similar hard attention was used in multiple object recognition~\cite{Ba-attention-2015}, object localization~\cite{Bellver_2016_NIPSWS,Mathe_2016_CVPR,NIPS2016_6532}, saliency map generation~\cite{Kuen_CVPR2015}, or action detection~\cite{Yeung_CVPR2016}.
While the early hard attention glimpses were not differentiable, implying reinforcement learning, the DRAW algorithm~\cite{Gregor2015_draw} and spatial transformer networks (STN)~\cite{Jaderberg2015-fu} provide attention crops which are fully differentiable and can thus be learned using gradient-based optimization.




Besides attention-based modules, the addition of external memory proved to increase the capacity of neural networks, by storing long-term information from past observations. This was mainly popularized by memory networks~\cite{Sukhbaatar_Memory_NIPS2015, Kumar_2016_ICML}. In \cite{Abdulnabi_2017_CVPR}, a Fully Convolutional Network is coupled with an attention-based memory
module to perform context selection and refinement, for semantic segmentation. In \cite{TokmakovSchmidArxiv2017}, visual memory is used to learn a spatio temporal representation of moving objects in a scene. The memory is implemented as a convolutional GRU with a 2D spatial hidden state. In \cite{Liu_GCA-LSTM_CVPR2017}, the ST-LSTM method of~\cite{Liu2016} is extended with a global context memory for skeleton-based action recognition. Multiple attention iterations are performed to optimize the global context memory, which is used for the final classification. In \cite{Sun_2017_ICCV}, an LSTM-based memory network is used for RGB and optical flow-based action recognition.

Our attention process is different from previously published work in that it produces an unstructured Glimpse Cloud in a spatio temporal cube. The attention process in unconstrained, which we show to be an important design choice. In our work, the external memory module is trainable in reading only, and provides a way to remember past soft-assignments of glimpses to recurrent workers.

\section{Dynamic sequential attention}

\noindent
We first introduce the following notation.
We want to map our input video sequence $\bX  \in \mathbb{R}^{T \times H \times W \times  3}$ to a corresponding activity label $y$ where $H$, $W$, $T$ denote, respectively, the height, the width and the number of time steps.
The sequence $\bX$ is a set of RGB input images $\bX_t \in \mathbb{R}^{H \times W \times  3}$ with $t=1...T$.
We do not assume any other kind of prior information on the input data.
We do not use any external information during testing such as pose data nor depth nor motion.
However, if pose data is available during \emph{training time}, our method is capable of integrating it in the form of additional inputs, which we show increases the performance of the system (see section \ref{sec:training}).


\noindent
Most of the RGB-only state-of-the-art methods, which do not use pose data, extract features at a frame level by feeding the entire video frame to a pre-trained deep network. This leads to global features, which do not capture local information well, which is relevant to the activities at hand.
Reasoning at a local level has, up till now, been achieved using pose features, or attention processes which were limited to attention maps (e.g.~\cite{Sharma2016a,videoLSTM}). Here, we propose an alternative approach, where an attention process runs statically over each time instant \emph{\textbf{and}} over time, creating sequences of sets of glimpse points, from which features are extracted.

Our model processes videos using several key components, also illustrated in figure \ref{fig:teaser}: i) a \emph{\textbf{recurrent spatial attention model}} that extracts features from different local glimpses $v_{t,g}$ following an attention path in each video over frames $t$ and multiple glimpses $g$ in each frame; and ii) \emph{\textbf{recurrent soft-tracking workers}} (indexed by $c$) which process these spatial features sequentially. The input data being unstructured, the spatial glimpses are soft-assigned to the workers, such that no hard decisions are taken at any point. To this end, iii) an
\emph{\textbf{external memory module}} keeps track of the glimpses seen in the past, their features, as well as of past soft-assignments, and produces new soft-assignments optimizing spatio-temporal consistency.
Our approach is fully-differentiable, such that the full model is trained end-to-end.

\subsection{Joint global/local feature space}
\label{sec:globallocal}
\noindent
We recognize activities based on global and local features jointly. In order to speed up calculations and to avoid extracting redundant calculations, we use a single feature space computed by a global model.
In particular, we map an input sequence $\bX$ to a spatio-temporal feature map $\bZ \in \mathbb{R}^{T \times H{'} \times W{'} \times  C{'}}$ using a deep neural network $f(\cdot)$ with 3D convolutions. Pooling is performed on the spatial dimensions but, not in time. This allows retention of the original temporal scale of the video, and thus access to features in each frame. It should, however, be noted, that due to the 3D convolutions used, the temporal receptive field of a single ``temporal'' slice of the feature map is greater than a single frame.   This is intended, as it allows the attention process to utilize motion. In an abuse of terminology, we will still use the term \emph{frame} to specify the slice $\bZ_t$ of a feature map with a temporal length of 1.
More information on the architecture of $f(\cdot)$ is given in section \ref{sec:IResnet3D}.


\subsection{A recurrent model of spatial attention}
\label{sec:spatialattention}
\noindent
Inspired by human behavior when scrutinizing a scene, we extract a fixed number of features from a series of $G$ glimpses within each frame. The process of moving from one glimpse to another is achieved with a recurrent model.
Glimpses are indexed by index $g{=}1\dots~G$, and each glimpse $\bZ_{t,g}$ corresponds to a sub-region of $\bZ_t$ using coordinates and scale $\bl_{t,g} = \left[   x_{g}, y_{g}, s_{g}^x,  s_{g}^y  \right ]^\top_t$ output by a differentiable glimpse function, which will be defined in section \ref{sec:draw}.  Features are extracted from the glimpse region $\bZ_{t,g}$ using global average pooling, resulting in a 1D feature vector $\bz_{t,g}$:
\begin{equation}
\bz_{t,g} = \Gamma (\bZ_{t,g}) = \frac{1}{H'W'}\sum_m \sum_n \bZ_{t,g}(m,n)
\label{eq:gap_Z_g}
\end{equation}
where $W' \times H'$ is the size of the glimpse region. The glimpse locations and scales $\bl_g$ for $g{=}1\dots~G$ are predicted by a recurrent network, which runs over glimpses. As illustrated in Fig.~\ref{fig:teaser}, the model predicts a fixed-length sequence of glimpse points for each frame.
It runs over the video, i.e.~it is not restarted/reinitialized after each frame. The hidden state thus carries information across frames and creates a globally coherent scrutinization process over the video. In equations \ref{eq:rnn_r} and \ref{eq:rnn_l} we index glimpses with a linear index $g$. The recurrent model is given as follows
(we use GRUs \cite{Cho2014-bh}, for notational simplicity we omit gates and biases in the rest of the equations):
\begin{gather}
\bh_{g} = \Omega (\bh_{g-1}, \left[ \bz_{g-1}, \br_t \right ] | \theta)  \label{eq:rnn_r} \\
\bl_{g} =  W^\top_l  \left[  \bh_{g}, \bc_t \right ] \label{eq:rnn_l}
\end{gather}
where $\bh$ denotes the hidden state of the RNN running over glimpses $g$, $\bc_t$ is a frame context vector for making the process aware of frame transitions (described in section \ref{sec:context}) and $\br_t$ carries information about the high level classification task. In essence, $\br_t$ corresponds to the global hidden state of the recurrent workers performing the actual recognition, as described in section \ref{sec:workers}, equation (\ref{eq:lstm}). Note, that the recurrence runs over glimpses $g$. The index $t$ here corresponds to the frame associated to current glimpse $g$.


\subsection{Differentiable glimpse module}
\label{sec:draw}
\noindent
In order to create a model which can be trained end-to-end, we use a simple version of spatial transformer networks (STN) \cite{Jaderberg2015-fu} to perform a differentiable crop operation on each feature map.
Given an input feature map $\bZ_t \in \mathbb{R}^{H \times W \times  C}$ and the glimpse parameters $\bl_g{=}\left[ x_g, y_g, s^x_g,  s^y_g \right]$ where $(x_g, y_g)$ is the central focus point and $(s^x_g,  s^y_g)$ corresponds to the scale, we output a feature map $\bZ_{t,g} \in \mathbb{R}^{H' \times W' \times  C}$. Note that the output size can differ from the input size.

We constrain the STN to implement a simple 2D affine transformation $A_{\bl_g}$ which allows cropping, translation and isotropic scaling on a regular grid point $x^t_i, y^t_i$ according to the given glimpse parameters $\bl_g$:

\begin{gather}
\left (
\begin{array}{c}
x^s_i \\
y^s_i \\
\end{array}
\right )
= A_{\bl_g}
\left (
\begin{array}{c}
x^t_i \\
y^t_i \\
1
\end{array}
\right )
=
\begin{bmatrix} s_g^x & 0 & x_g \\ 0 & s_g^y & y_g \end{bmatrix}
\left (
\begin{array}{c}
x^t_i \\
y^t_i \\
1
\end{array}
\right )   \label{eq:stn}
\end{gather}
where $x^t_i, y^t_i$ are the target coordinates of the regular grid in the output feature map $\bZ_{t,g}$ and $x^s_i, y^s_i$ are the source coordinates in the input feature map that define the sample points.

We must define a sampler which takes the set of sampling points $(x^s_i, y^s_i)$, along with the input feature map $\bZ_t$ and produces the sampled output feature map $\bZ_{t,g}$. We employ bilinear interpolation which implements the following mapping:
$$
\begin{array}{l}
\bZ_{t,g}(x^s_i, y^s_i) = \\
\sum_{n}^{H'} \sum_{m}^{W'} \bZ_t(m,n) \max (0, 1{-}|x^s_i{-}n|) \max (0, 1{-}|y^s_i{-}m|) .\
\end{array}
$$
The STN is differentiable, which allows us to train the parameters $W_l$ for the prediction of focus point parameters $\bl_g$ together with the rest of the network using gradient descent.


\section{Distributed Reasoning on Unstructured Glimpse Clouds}

\begin{figure}[t] \centering
    \centering
        \includegraphics[width=8cm]{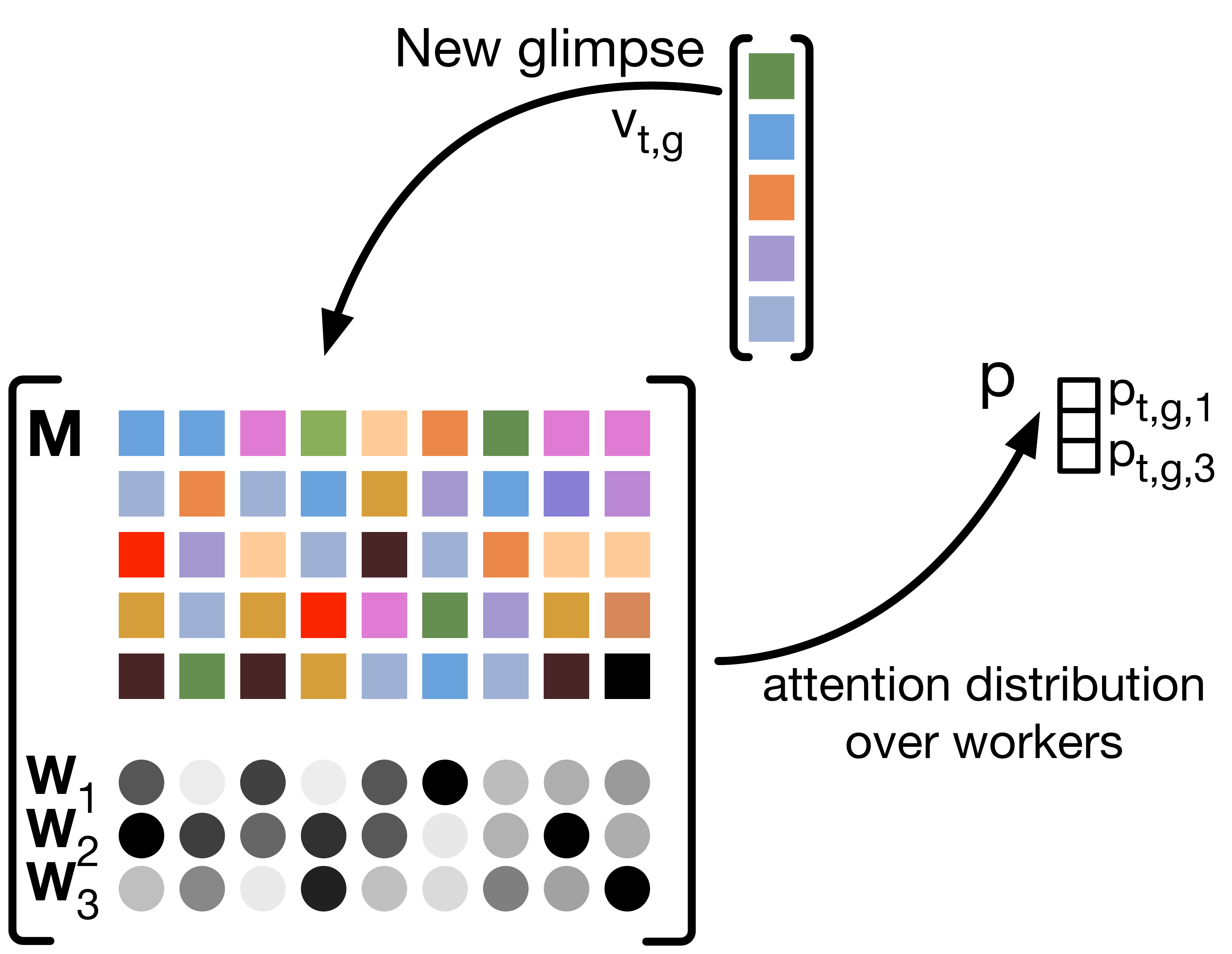}
    \caption{An external memory module determines an attention distribution over workers (a soft assignment) for each new glimpse $\bv_{t,g}$ based on similarities with past glimpses $M$ and their past attention probabilities $w$. Shown for a single glimpse and 3 workers.}
    \label{fig:mn}
\end{figure}

\noindent
The attended points (glimpses) predicted in each frame $\bZ_t$
have a semantic meaning in the input video (e.g.~a patch around the hands or shoulders; an object held or pointed at by a person etc.). The goal is to reason about their positions, motion, changes in appearance, relationships or other properties.
This is made difficult by the
sequential attention process described in section \ref{sec:spatialattention}, which
can provide very different glimpse sequences for each frame, since we avoid any direct supervision. This is intentional, in order to give the spatial attention process complete freedom. In particular, it can choose to jump to different glimpse points at each frame, and/or decide to revisit certain glimpses attended to in the past. Since frame features $\bZ_t$ also encode motion due to the 3D convolutions in $f(\cdot)$, the attention process can learn to revisit attended points, compensating for their motion. In section \ref{sec:experiments} we describe experiments performed which justify this kind of attention process compared to an alternate choice of spatio-temporally coherent attention.

As a consequence, extracting motion cues from semantic points in the scene requires associating glimpse points from different frames over time. Due to the freedom of the attention process and fixed number of glimpses, subsequent glimpses of the same point in the scene are generally not in subsequent frames, which excludes conventional tracking mechanisms known from the computer vision literature. Instead, we
avoid hard tracking and hard assignments between glimpse points in a temporal manner.
We propose a soft associative model for automatically associating similar spatial features over time.


\subsection{Distributed soft-tracking workers}
\label{sec:workers}
\noindent
As given in eq.~(\ref{eq:gap_Z_g}), we denote by $\bz_{t,g}$ the features extracted from the $g^{th}$ glimpse in  feature map  $\bZ_t$ for $g=1...G$ and $t=1...T$.
We are interested in a joint encoding of spatial dimensions and feature dimensions and employ
``\emph{what}'' and ``\emph{where}'' features $\bv_{t,g}$ introduced in \cite{Larochelle_NIPS2010} defined by:
\begin{gather}
\bv_{t,g} = \bz_{t,g} \otimes \Lambda (\bl_{t,g} | \theta_{\Lambda})
\end{gather}
where $\otimes$ is the Hadamard product and $ \Lambda (\bl_{t,g} | \theta_{\Lambda})$ is a network which provides an embedding of the spatial patch coordinates into a space which is of the same dimensionality as the features $\bz_{t,g}$.
The vector $\bv_{t,g}$ contains joint cues about motion and appearance, but also the spatial localization of those features.

Evolution over time of this information is modeled with a number ($C$) of so-called \emph{soft-tracking workers} $\Psi_{c}$ for $c=1...C$.
Each worker corresponds to a recurrent model capable of tracking entities over time.
We \emph{never} hard assign glimpses to workers.
Inputs to each individual worker correspond to weighted contributions from all of the $G$ glimpses. In general, the number of glimpse points $G$ can be different from the number of workers $C$.
At each time instant, focal points are thus soft-assigned to the workers on the fly but changing the weights of the contributions, which will be described further below.

A worker $\Psi_{c}$ is a recurrent network following the usual update equations based on the past state $\br_{t-1,c}$ and its input $\tilde{\bv}_{t,c}$:
\begin{gather}
\br_{t,c} = \Psi_{c} (\br_{t-1,c}, \tilde{\bv}_{t,c} | \theta_{\Psi_{c}}) \\
\br_t = \sum_c \br_{t,c}
\label{eq:lstm}
\end{gather}
where $\Psi_{c}$ is a GRU and $\br_t$ is carrying global information about the current state (needed as input of the recurrent model of spatial attention).
The input $\tilde{\bv}_{t,c}$ to each worker $\Psi_{c}$ is a linear combination of the different glimpses $\{ \bv_{t,g} \}, g=1\dots~G$ weighted by a soft attention distribution $\bp_{t,c} = \{ p_{t,g,c} \}, \ g=1\dots~G$:
\begin{equation}
\tilde{\bv}_{t,c} = \bV_{t} \bp_{t,c}
\label{eq:inputweighting}
\end{equation}
where $\bV_t$ is a matrix whose rows are the different glimpse features $\bv_{t,g}$.
Workers are independent from each other in the sense that they do not share parameters $\theta_{\Psi_{c}}$.
This can potentially lead to specialization of the workers on types of tracked and integrated scene entities.

\subsection{Soft-assignment using External Memory}
\noindent
The role of the attention distribution $\bp_{t,c}$ is to give higher weights to glimpses which have been soft-assigned to this worker in the past.
Thus workers extract different kinds of features from each other.
To do so, we employ an external memory bank denoted $\bM = \{ \bm_k \}$ which is common to all workers.
In particular, $\bM$ is a fixed-length array of $K$ entries $\bm_k$ each capable of storing a feature vector $\bv_{t,g}$.
Even if the external memory is common to each worker, they have their own ability to extract information from it.
Each worker $\Psi_{c}$ has its own weight bank denoted $\bW_c= \{ w_{c,k} \}$.
The scalar $w_{c,k}$ holds the importance of the entry $\bm_{c,k}$ for worker $\Psi_{c}$ .
Hence the overall external memory is defined by the set $\{\bM,\bW_1, \dots \bW_c \}$.

\paragraph{Attention from memory reads} ---
The attention distribution $\bp_{t,c}$ is a distribution over glimpses $g$, i.e. $\bp_{t,c} = \{ p_{t,c,g} \}$, $ 0 \leq p_{t,c,g} \leq 1$ and $ \sum_g p_{t,c,g}{=}1.$
We want the glimpses to get distributed appropriately across the workers, and encourage worker specialization.
In particular, at each timestep we want to assign a glimpse high importance to a worker if this worker has been soft-assigned similar glimpses in the past with high importance. To this end, we define a fully trainable distance function $\phi(.,.)$ which is implemented as a quadratic form:
\begin{gather}
	\phi(\bx,\by) = \sqrt{(\bx-\by)^\top \bD (\bx-\by)} \label{eq:mahalanobis}
\end{gather}
where $\bD$ is a learned weight matrix.
Within each batch we normalize $\phi(\cdot,\cdot)$ by min-max normalization to scale it to lie between 0 and 1.

A glimpse $g$ is soft-assigned to a given worker $c$ with a higher weight $p_{t,c,g}$ if $\bv_{t,g}$ is similar to vectors $\bm_{k}$ from the memory bank $\bM$ which had a high importance for the worker in the past $\Psi_{c}$ :
\begin{gather}
p_{t,c,g} =  \sigma_\alpha \left ( \sum_k e^{-t^{m_k}} \times w_{c,k}
	\left [
	1-  \phi(\bv_{t,g}, \bm_{k})
	\right ]
\right )
\end{gather}
where $\sigma$ is the softmax function over the $G$ glimpses and $e^{-t^{m_k}}$ is an exponential rate over time to give higher importance to recent feature vectors compared to those in the distant past.
$t^{m_k}$ is the corresponding timestep of the memory bank $m_k$.
In practice we add a temperature term $\alpha$ to the softmax function $\sigma$. When $\alpha \to 0 $ the output vector is sparser. The negative factor multiplied with $\phi$ is justified by the fact that $\phi$ is initially pre-trained as a Mahalanobis distance by setting $\bD$ to the inverse covariance matrix of the glimpse data. The factor therefore transforms the distance into a similarity. After pre-training, $\bD$ is trained end-to-end.

The attention distribution $\bp_{t,c}$ is computed for each worker $\Psi_{c}$.
Thus each glimpse $g$ potentially contributes to each worker $\Psi_{c}$ through its input vector $\tilde{\bv}_{t,c}$ (c.f.~equation (\ref{eq:inputweighting})), albeit with different weights.


\paragraph{Memory writes ---} for each frame,
the feature representations $\bv_{t,g}$ are stored in the memory bank $\bM$.
However, the attention distribution $\bp_{t,c} = \{ p_{t,c,g} \}$ is used to weight these entries for each worker $\Psi_{c}$.
If a glimpse feature $\bv_{t,g}$ is stored in a slot $\bm_{k}$, then its importance weight  $w_{c,k}$ for worker $\Psi_{c}$ is set to $p_{t,c,g}$.
The only limitation is the size $K$ of the memory bank. When the memory is full,
we delete the oldest memory slot.
More flexible storing processes, e.g.~trained mappings, are left for future work.

\subsection{Recognition}
\label{sec:recognition}
\noindent
Since workers proceed in a independent manner through time, we need an aggregation strategy to perform classification.
Each worker $\Psi_{c}$ has its own hidden state $\left\lbrace \br_{t,c} \right\rbrace _{t=1...T}  $ and is responsible for its own classification through a fully-connected layer.
The final classification is done by averaging logits of the workers:
\begin{gather}
	\bq_c =W_c \cdot \br_c \\
	\hat{\by} =  \text{softmax} \left(
	\sum_{c}^{C} \bq_c
	\right)
\end{gather}
where $\hat{\by}$ is the probability vector of assigning the input video $\bX$ to each class.

\subsection{Context vector}
\label{sec:context}
\noindent
In order to make the spatial attention process (section \ref{sec:spatialattention}) aware of frame transitions, we introduce a context vector $\bc_t$ which contains high level information about humans present in the current frame $t$.
$\bc_t$ is obtained by global average pooling over the spatial domain of the penultimate feature maps of a
given timestep.
We regress the 2D pose coordinates of humans from the context vector $\bc_t$ using the following mapping:
\begin{gather}
\by^p_t = W_p^\top \bc_t
\end{gather}
Pose $\by^p_t$ is linked to ground truth pose (during \emph{training} only) using a supervised term described in section \ref{sec:training}. This leads to incorporate  hierarchical feature learning in a sense that the penultimate feature maps have to detect human joints present in each frame.

\section{Training}
\label{sec:training}
\noindent
We train the model end-to-end with a sum of a collection of loss terms, which are explained in the following paragraphs:
\begin{equation}
\mathcal{L} = \mathcal{L}_D (\hat{\by},\by) + \mathcal{L}_P(\hat{\by}^p,\by^p) +  \mathcal{L}_G(\bl,\by^p)
\end{equation}

\myparagraph{Supervision}  $\mathcal{L}_D (\hat{\by},\by)$ is a supervised loss term (cross-entropy loss on activity labels $\by$).

\myparagraph{Pose prediction} articulated pose $\by^p$ is available for many datasets. Our goal is to \emph{not} depend on pose during testing; however, its usage during training can provide additional information to the learning process and reduce the tendency of activity recognition methods to memorize individual elements in the data for recognition. We therefore add an additional term
$\mathcal{L}_P(\hat{\by}^p,\by^p)$, which encourages the model to perform pose regression during training only from intermediate feature maps (described in section \ref{sec:context}).
Pose regression over time leads to a faster convergence of the overall model.

\myparagraph{Attracting glimpses to humans}
$\mathcal{L}_G({\bl},\by^p)$ is a loss encouraging the glimpse points to be as sparse as possible within a frame but by the same time close to humans in the scene.
Recall that $\bl_{t,g} =\left[   x_{t,g}, y_{t,g}, s_{t,g}^x,  s_{t,g}^y \right ]^T$, so
$\mathcal{L}_G$ is defined by:
\begin{gather}
	\mathcal{L}^t_{G_1}(\bl,\by^p) = \frac{1}{1 +\sum_{g_1}^{G} \sum_{g_2}^{G} || \bl_{t,g_1},  \bl_{t,g_2} || }\\
	\mathcal{L}^t_{G_2}(\bl,\by^p) = \sum_{g}^G \min_{j} || \bl_{t,g}, \by^p_j || \\
	\mathcal{L}_G(\bl,\by^p) = \sum_{t}^T \left( \mathcal{L}^t_{G_1}(\bl,\by^p) + \mathcal{L}^t_{G_2}(\bl,\by^p)\right)
\end{gather}
where $\by^p_j$ denotes the 2D coordinates of joints $j$, and Euclidean distance on $\bl_{t,g}$ is computed using the central focus point $(x_{t,g},y_{t,g})$ only.
$\mathcal{L}_{G_1}$ encourages diversity between glimpses within a frame.
$\mathcal{L}_{G_2}$ ensures that all the glimpses are not taken too far away from humans.

\section{Architectures - Pretraining}
\label{sec:IResnet3D}
\noindent
We designed the 3D convolutional network $f(\cdot)$ computing the global feature maps in section \ref{sec:globallocal} such that the temporal dimension is maintained, i.e. without any temporal subsampling.
We proceed from the Resnet-50 network\cite{He2015} and inflate the 2D spatial convolutional kernels into 3D kernels, artificially creating new a temporal dimension, as described by Carreira et al \cite{Carreira_2017_CVPR}. This allows us to take advantage of the 2D kernels learned by pre-training on image classification on the Imagenet dataset.
The Inflated ResNet $f(\cdot)$ is then trained as a first step by minimizing the loss $\mathcal{L}_D + \mathcal{L}_P$.
The supervised loss $\mathcal{L}_D $ on the global model is applied on a path attached to global average pooling on the last feature maps followed by a fully-connected layer which is subsequently removed.

The recurrent spatial attention module $\Omega(\cdot)$ is a GRU with a hidden state of size 1024;  $\Lambda(\cdot)$ is an MLP with a single hidden layer of size 256 and a ReLU activation; the soft-trackers $\Psi_{c}$ are GRUs with a hidden state of size 512. There is no parameter sharing between them.

\section{Experimental results}
\label{sec:experiments}

\begin{figure*}[t]
    \centering
        \includegraphics[width=4cm,height=3cm]{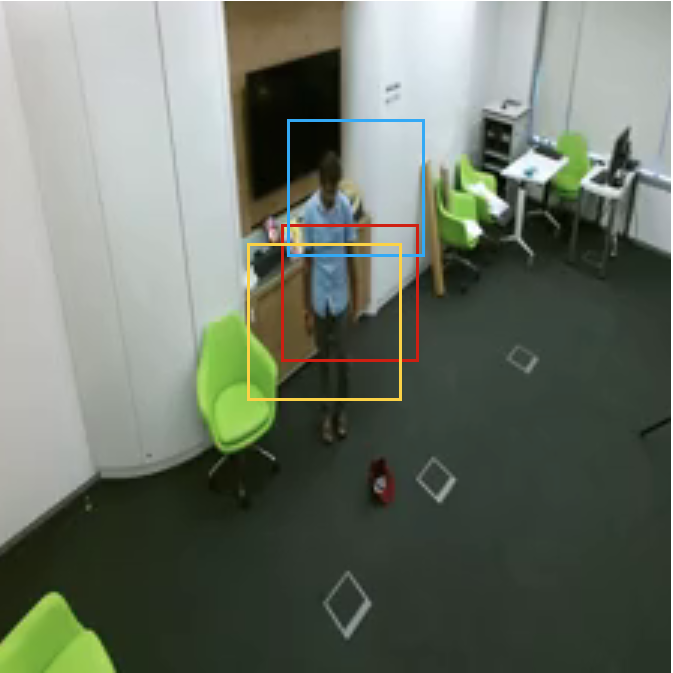}
        \includegraphics[width=4cm,height=3cm]{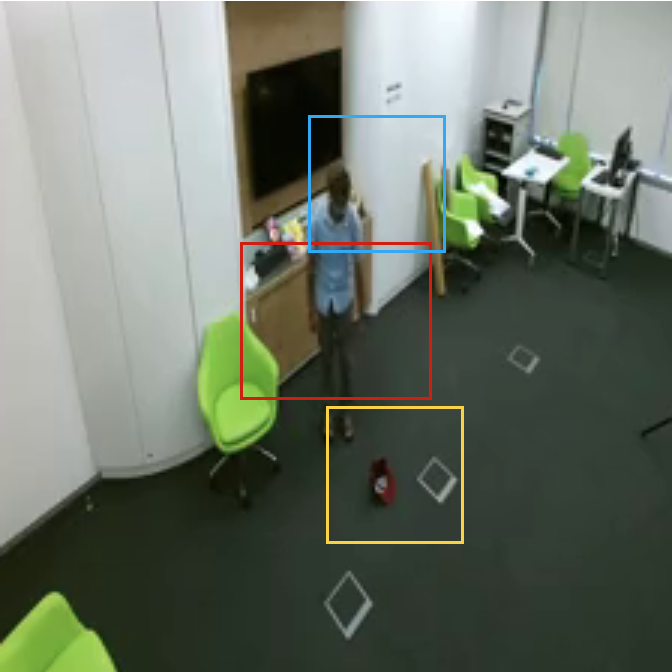}
        \includegraphics[width=4cm,height=3cm]{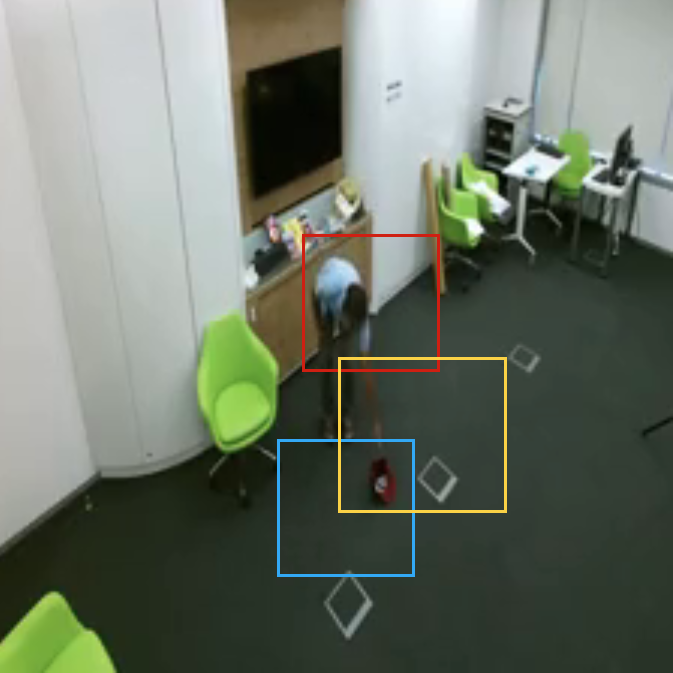}
        \includegraphics[width=4cm,height=3cm]{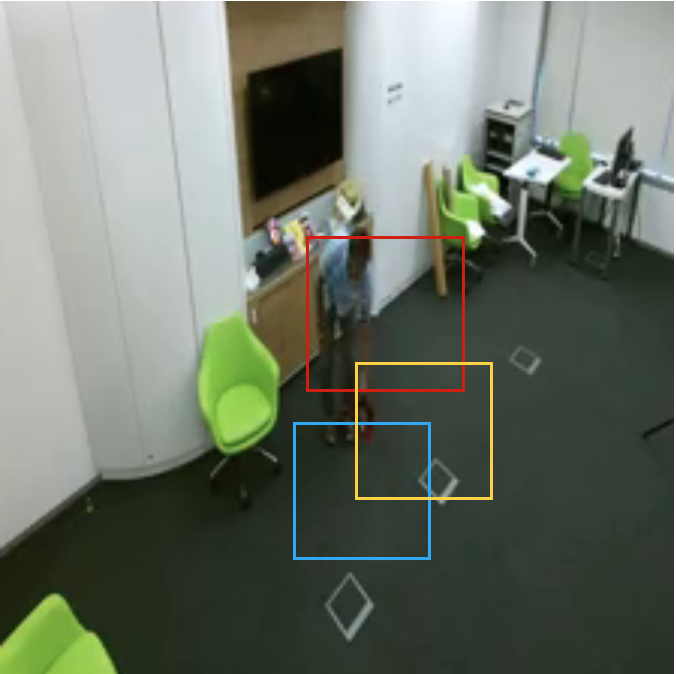}

		\includegraphics[width=4cm,height=3cm]{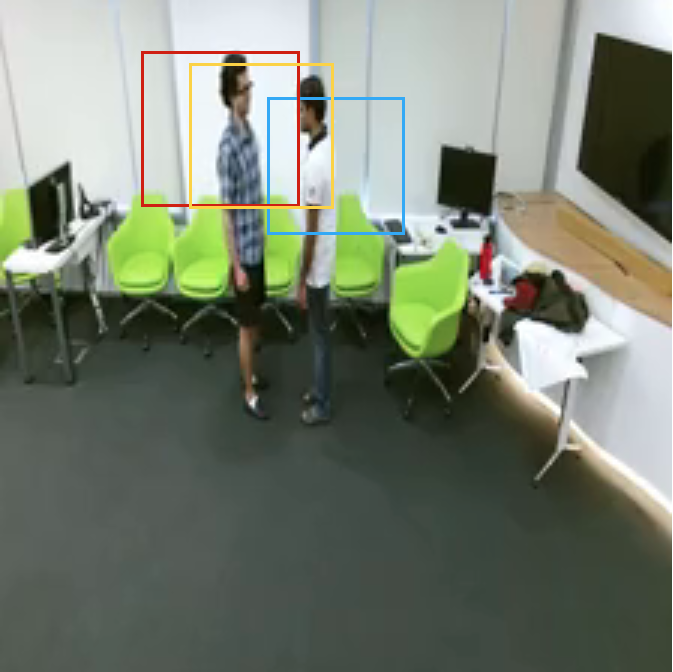}
		\includegraphics[width=4cm,height=3cm]{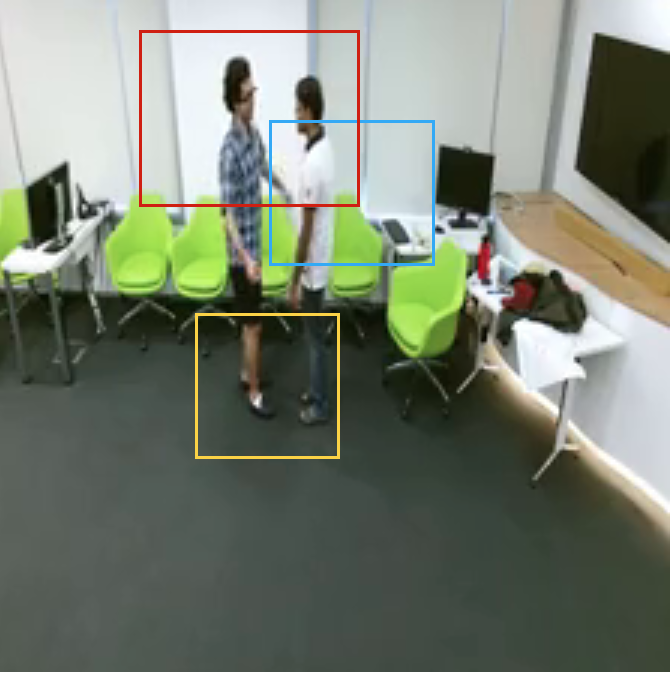}
		\includegraphics[width=4cm,height=3cm]{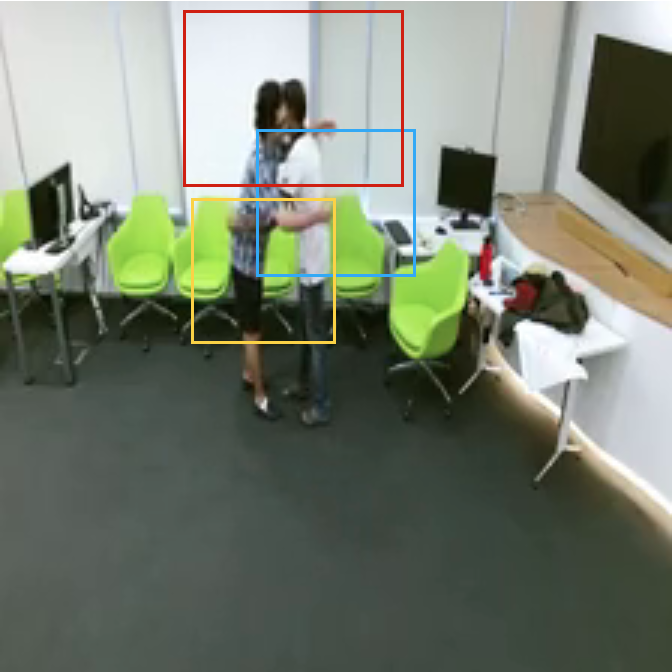}
		\includegraphics[width=4cm,height=3cm]{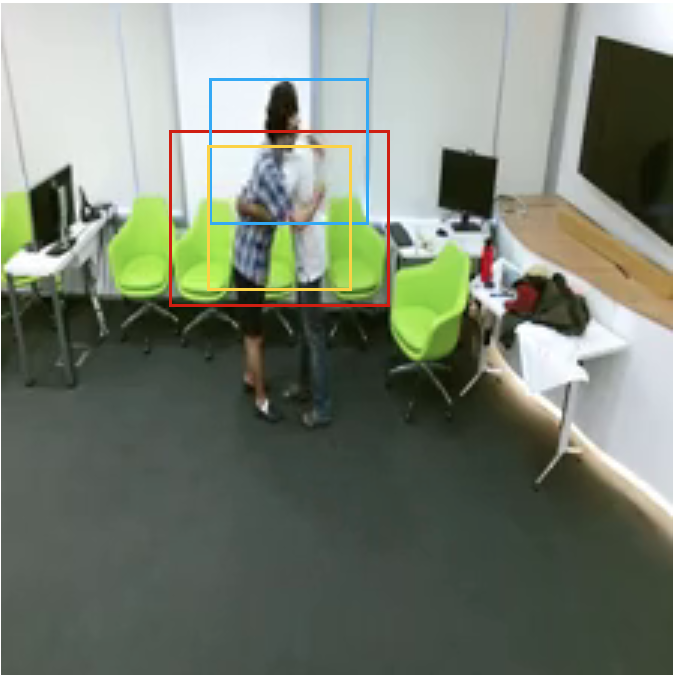}
    \caption{An illustration of the glimpse distribution for several sequences of the NTU dataset. Here we set 3 glimpses per frame (G=3, Red: first glimpse, Blue: second glimpse, Yellow: third one).}
    \label{fig:glimpsedistributions}
\end{figure*}

\noindent

\noindent
The proposed method has been evaluated on two human action recognition datasets: NTU RDB+D Dataset~\cite{Shahroudy2016} and Northwestern-UCLA Multiview Action 3D Dataset~\cite{Wang_2014_CVPR}.
\myparagraph{NTU RDB+D Dataset (NTU)}
NTU has been acquired with a Kinect v2 sensor and contains more than 56K videos and 4 million frames with 60 different activities including individual activities, interactions between 2 people and health related events. The actions have been performed by 40 subjects and from 80 viewpoints.
We follow the cross-subject and cross-view split protocol from~\cite{Shahroudy2016}.
Due to the large amount of videos, this dataset is highly suitable for deep learning modeling.

\myparagraph{Northwestern-UCLA Multiview Action 3D Dataset (N-UCLA)}
This dataset \cite{Wang_2014_CVPR} contains 1494 sequences, covering ten action categories, such as \textit{drop trash} or \textit{sit down}.
Each sequence is captured simultaneously by 3 Kinect v1 cameras.
RGB, depth, and human pose are available for each video, and each action is performed one to six times by ten different subjects.
Most actions involve human-object interaction, making this dataset challenging.
We followed the cross-view protocol defined by \cite{Wang_2014_CVPR}, and we trained our method on samples from two camera views, and tested it on samples from the remaining view.
This produced three possible cross-view combinations: $V_{1,2}^3, V_{1,3}^2, V_{2,3}^1$.
The combination $V_{1,2}^3$ means that samples from view 1 and 2 are used for training, and samples from view 3 are used for testing.

%


\myparagraph{Implementation details}
Following \cite{Shahroudy2016}, we cut videos into sub-sequences of 8 frames and sample sub-sequences. During training, a single sub-sequence is sampled. During testing, 5 sub-sequences and logits are averaged.
RGB videos are rescaled to $256\times256$ and random cropping of size $224\times224$ is done during training and testing.

Training is performed using the Adam Optimizer \cite{AdamOptimization2015} with an initial learning rate of 0.0001. We use minibatches of size 40 on 4 GPUs.
 Following \cite{Shahroudy2016}, we sample 5\% of the initial training set as a validation set, which is used for hyper-parameter optimization and for early stopping.
 All hyperparameters have been optimized on the validation sets of the respective datasets.
 We used the model trained on NTU as a pre-trained model and fine-tuned it on N-UCLA.


\myparagraph{Comparison with the state of the art}
Our method outperforms state-of-the-art methods on NTU and N-UCLA by a large margin, and this also includes several methods which use multiple modalities among RGB, depth and pose.
Table \ref{table:NTUSOTA} and \ref{table:N-UCLA_SOTA} provide detailed results compared to the state-of-the-art on the NTU dataset.

\begin{table}
	\begin{center}
		\caption{Results on the Northwestern-UCLA Multiview Action 3D dataset with Cross-View Setting (accuracy as a percent). V, D, and P mean Visual (RGB), Depth, and Pose, respectively.}
		\begin{tabular}{cccccc}
			\arrayrulecolor{tablecolor} \toprule
			Methods                                 & {\footnotesize Data} & $V_{1,2}^3$ & $V_{1,3}^2$ & $V_{2,3}^1$ & Avg \\
            \arrayrulecolor{tablecolor} \toprule
			DVV \cite{Li_2014_CVPR}         		& D 	& 58.5 			& 55.2 		& 39.3 		& 51.0  \\
			CVP \cite{Zhang_2013_CVPR}      		& D 	& 60.6 			& 55.8 		& 39.5 		& 52.0  \\
			AOG \cite{Wang_2014_CVPR}       		& D 	& 45.2 			& -    		& -    		& -     \\
			HPM+TM \cite{Rahmani_2016_CVPR} 		& D 	& \textbf{91.9}	& 75.2 		& 71.9 		& 79.7  \\	\hline
			Lie group \cite{Vemulapalli_2014_CVPR}  & P 	& 74.2 			& -    		& -    		& -     \\
			HBRNN-L \cite{Du_2015_CVPR}		 	   	& P 	& 78.5 			& -    		& -    		& -     \\
			Enhanced viz. \cite{LIU2017346}	 	   	& P 	& 86.1 			& -    		& -    		& -     \\
			Ensemble TS-LSTM \cite{Lee_2017_ICCV}  	& P 	& 89.2 			& -   		& -   		& -     \\	\hline
			Hankelets \cite{Li_2012_CVPR}   	   	& V 	& 45.2 			& -   		& -   		& -     \\
			nCTE \cite{Gupta_2012_CVPR}     	   	& V   	& 68.6 			& 68.3 		& 52.1 		& 63.0  \\
			NKTM \cite{Rahmani_2015_CVPR}   		& V   	& 75.8 			& 73.3 		& 59.1 		& 69.4  \\ \hline \hline
			\textbf{Global model}    						& V     & 85.6 & 84.7 & 79.2 & 83.2\\
            \textbf{Glimpse Clouds	}    						& V     & 90.1 & \textbf{89.5} & \textbf{83.4} & \textbf{87.6}\\
            \arrayrulecolor{tablecolor} \bottomrule
		\end{tabular}
	\end{center}
	\label{table:N-UCLA_SOTA}
\end{table}

\begin{table}
	\begin{center}
		\begin{tabular}{cccccc}
			\arrayrulecolor{tablecolor} \toprule
			Methods                                  & {\footnotesize Pose} &  {\footnotesize RGB} & CS & CV & Avg \\
            \arrayrulecolor{tablecolor} \toprule
			Lie Group \cite{Vemulapalli_2014_CVPR}     & \checkmark & -& 50.1          & 52.8       & 51.5    \\
			Skeleton Quads \cite{Evangelidis-ICPR-2014} & \checkmark & -&38.6          & 41.4       & 40.0    \\
			Dynamic Skeletons \cite{Hu_CVPR2015}      & \checkmark & -&60.2          & 65.2       & 62.7    \\
			HBRNN \cite{Du_CVPR2015}     & \checkmark & -&59.1          & 64.0       & 61.6    \\
			Deep LSTM \cite{Shahroudy2016}        & \checkmark & -&60.7          & 67.3       & 64.0    \\
			Part-aware LSTM \cite{Shahroudy2016}  & \checkmark & -&62.9          & 70.3       & 66.6    \\
			ST-LSTM + TrustG.  \cite{Liu2016}    & \checkmark & -&69.2          & 77.7       & 73.5    \\
			STA-LSTM   \cite{Song2016}         & \checkmark & -&73.2          & 81.2       & 77.2    \\
			Ensemble TS-LSTM \cite{Lee_2017_ICCV} & \checkmark & -&74.6          & 81.3       & 78.0    \\
			GCA-LSTM   \cite{Liu_GCA-LSTM_CVPR2017}         &  \checkmark & -&74.4          & 82.8      & 78.6    \\
			JTM   \cite{Wang}          & \checkmark & -& 76.3   & 81.1   &  78.7 \\
			MTLN  \cite{Ke_2017_CVPR}          & \checkmark & -& 79.6  & 84.8   &  82.2 \\
			VA-LSTM   \cite{Zhang_2017_ICCV}          & \checkmark & -& 79.4   & 87.6   & 83.5  \\
			View-invariant   \cite{LIU2017346}          & \checkmark & -& 80.0   & 87.2   & 83.6  \\  \hline
            DSSCA - SSLM \cite{Shahroudy20162} & \checkmark & \checkmark & 74.9 & - & -  \\
            STA-Hands \cite{Baradel_2017_ICCV_Workshops} & X & X& 82.5 & 88.6 & 85.6   \\
            Hands Attention \cite{BaradelArxiv2017HandAttention} &  \checkmark & \checkmark& 84.8       &   90.6 & 87.7   \\ \hline
            C3D$\dagger$  & - & \checkmark& 63.5       &  70.3 &  66.9  \\
            Resnet50+LSTM$\dagger$   & - & \checkmark& 71.3       &  80.2 & 75.8   \\ \hline \hline
            \textbf{Glimpse Clouds}    & - & \checkmark& \textbf{86.6}       &   \textbf{93.2} & \textbf{89.9}   \\
            \arrayrulecolor{tablecolor} \bottomrule
		\end{tabular}
	\end{center}
	\caption{Results on the NTU RGB+D dataset with Cross-Subject and Cross-View settings (accuracies in \%);
	($\dagger$ indicates method has been re-implemented).}
	\label{table:NTUSOTA}
\end{table}

\begin{table*}[t]
    \begin{center}
        \begin{tabular}{ccccccccc}
            \arrayrulecolor{tablecolor} \toprule
           Methods     & Spatial Attention & Soft Workers & $L_D$ & $L_P$ & $L_G$ & CS& CV & Avg \\
            \arrayrulecolor{tablecolor} \toprule
            Global model   & -& -&  \checkmark & -& - & 84.5        & 91.5     & 88.0   \\\arrayrulecolor{tablecolor}
            Global model   & -& -&  \checkmark & \checkmark & - & 85.5        & 92.1     & 88.8   \\\arrayrulecolor{tablecolor}
            Global model+$\sum$ Glimpses + GRU   & -& -&  \checkmark & \checkmark & - & 85.8        & 92.4     & 89.1   \\\arrayrulecolor{tablecolor}
            Glimpse Clouds & \checkmark & \checkmark &  \checkmark & - & - &    85.7       &   92.5 & 89.1 \\\arrayrulecolor{tablecolor}
            Glimpse Clouds & \checkmark & \checkmark &  \checkmark & \checkmark & - &    86.4       &   93.0 & 89.7 \\\arrayrulecolor{tablecolor}
            Glimpse Clouds & \checkmark & \checkmark &  \checkmark & - & \checkmark &    86.1      &   92.9 & 89.5 \\\arrayrulecolor{tablecolor}
            Glimpse Clouds & \checkmark & \checkmark &  \checkmark & \checkmark & \checkmark &    \textbf{86.6}       &   \textbf{93.2} & \textbf{89.9} \\\arrayrulecolor{tablecolor}
            Glimpse Clouds + Global model & \checkmark & \checkmark &  \checkmark & \checkmark & \checkmark &    86.6       &   93.2 & 89.9 \\\arrayrulecolor{tablecolor}
            \bottomrule
        \end{tabular}
    \end{center}
    \caption{Results on NTU: ablation study}
        \label{table:ablation}
\end{table*}

\begin{table}
	\begin{center}
		\begin{tabular}{ccccc}
			\arrayrulecolor{tablecolor} \toprule
			Glimpse & Type of attention  & CS & CV & Avg \\
            \arrayrulecolor{tablecolor} \toprule
            3D tubes & Attention 		 & 85.8       &   92.7 & 89.2   \\
            \midrule
            Seq. 2D & Random sampling    & 80.3       &   87.8 & 84.0  \\
            Seq. 2D & Saliency           & 86.2       &   92.9 & 89.5   \\
            Seq. 2D & \textbf{Attention}          & \textbf{86.6}       &   \textbf{93.2} & \textbf{89.9}   \\
            \arrayrulecolor{tablecolor} \bottomrule
		\end{tabular}
	\end{center}
	\caption{Results on the NTU: different attention and alternative strategies.}
	\label{table:attentionstrategies}
\end{table}

\myparagraph{Ablation study}
Table \ref{table:ablation} shows several experiments to study the effect of our design choices. Classification from the Global Model alone (Inflated-Resnet-50) is clearly inferior to the distributed recognition strategy using the set of workers (+1.9 points on NTU and +4.4 points on N-UCLA).
The bigger gap obtained on N-UCLA can be explained by the larger portion of the frame occupied by people and therefore higher efficiency of a local representation.
The additional loss predicting pose during training helps, even though pose is not used during testing. An important question is whether the Glimpse Cloud could be integrated with an easier mechanism than a soft-assignment. We tested a baseline which sums glimpse features for each time step and which integrates them temporally (row \#3). This gave only a very small improvement over the global model. Distributed recognition from Glimpse Clouds with soft-assignment clearly outperforms the simpler baselines. Adding the global model does not gain any improvement.

\myparagraph{Importance of losses}
Table \ref{table:ablation} also shows importances of our three loss functions.
Cross-entropy only $L_D$ gives 89.1\%. Adding pose prediction $L_P$ we gain 0.6 points and adding pose attraction $L_G$ we gain 0.4 points, which are complementary.

\myparagraph{Unstructured vs. coherent attention}
we also evaluated the choice of unstructured attention, i.e. the decision to give the attention process complete freedom to attend to a new (and possibly unrelated) set of scene points in each frame. We compared this with an alternative choice, where glimpses are axis-aligned space-time tubes over the whole temporal length of the video. In this baseline, the attention process is not aligned with time. At each iteration, a new tube is attended in the full space-time volume, and no tracking or soft-assignment to worker modules is necessary. As indicated in Table~\ref{table:attentionstrategies}, this choice is sub-optimal. We conjecture that tubes cannot cope with moving objects and object parts in the video.

\myparagraph{Attention vs.~saliency vs.~random}
we evaluated whether a sequential attention process contributes to performance, or whether the gain is solely explained from the sampling of local features in the space-time volume. We compared our choice with two simpler baselines: (i) complete random sampling of local features, which leads to a drop of more than 6 points. The location of the glimpses is clearly important. (ii) with a saliency model, which predicts glimpse locations in parallel through different outputs of the location network. This is not a full attention process, in that a glimpse prediction does not depend on what the model has seen in the past. This choice is also sub-optimal.

\myparagraph{Learned weight matrix} Random initialization and fine-tuning  of $D$ matrix in equation \ref{eq:mahalanobis} loses 0.4 points and leads to slower convergence by a factor of 1.5. Fixing $D$ (to inverse covariance) w/o any training loses 0.8 points.

\myparagraph{The Joint encoding} (``what and where'' features) are important in order to correctly weight their respective contribution. Plainly adding concatenating coordinates and features loses 1.1 points.

\myparagraph{Hyper-parameters $C$, $G$, $T$} Number of glimpses and workers: $C$ and $G$ were selected by cross-validation on the validation set by varying them from 1 to 4, giving an optimum of $G{=}C{=}3$ over all 16 combinations. More leads the model to overfit.
The size of the memory bank $K$ is set to $T$ where $T{=}8$ is the length of the sequence.

\myparagraph{Runtime} the model has been trained on a GPU cluster with a single job spread over 4 Titan Xp GPUs. Pre-training the global model on the NTU dataset takes 16h. Training the Glimpse Cloud model end-to-end then takes a further 12h. A single forward pass over the full model takes 97ms on 1 GPU. The method has been implemented in PyTorch.


\section{Conclusion}
\label{sec:conclusion}
\noindent
We proposed a method for human activity recognition which does not rely on depth images or articulated pose, though it is able to leverage pose information during training. The method achieves state-of-the-art performance on the NTU and N-UCLA datasets even when compared to methods that use pose, depth or both at test time. An attention process over space and time produces an unstructured \emph{Glimpse Cloud}, which are soft-assigned to a set of tracking/recognition workers. In or experiments, we showed that this distributed recognition outperforms a global convolutional model and also local models with simpler baselines for the localization of glimpses. Future work will investigate more complex procedures for fusing the decisions of the set of workers.

 \section{Acknowledgements}
This work was funded under grant ANR Deepvision (ANR- 15-CE23-0029), a joint French/Canadian call by ANR and NSERC.

{\small
\bibliographystyle{ieee}
\bibliography{refs}
}

\end{document}